\documentclass[runningheads,a4paper]{llncs}

\usepackage{amssymb}
\usepackage{booktabs}
\usepackage{graphicx}
\usepackage{varwidth}
\usepackage{caption}

\usepackage{url}
\urldef{\mail}\path|{ycui,rtobossi,ovigouroux}@gmf.fr|  
\newcommand{\keywords}[1]{\par\addvspace\baselineskip
\noindent\keywordname\enspace\ignorespaces#1}

\begin{document}

\mainmatter  

\title{Modelling customer online behaviours with neural networks: applications to conversion prediction and advertising retargeting}
\titlerunning{Modelling customer online behaviours with neural networks}

\author{Yanwei Cui \and Rogatien Tobossi \and Olivia Vigouroux}
\authorrunning{Modelling customer online behaviours with neural networks}

\institute{GMF Assurances, Groupe Cov\'{e}a \\
148 Rue Anatole France, 92300 Levallois-Perret, France \\
\mail\\
}

\maketitle

\begin{abstract}
In this paper, we apply neural networks into digital marketing world for the purpose of better targeting the potential customers. To do so, we model the customer online behaviours using dedicated neural network architectures. Starting from user searched keywords in a search engine to the landing page and different following pages, until the user left the site, we model the whole visited journey with a Recurrent Neural Network (RNN), together with Convolution Neural Networks (CNN) that can take into account of the semantic meaning of user searched keywords and different visited page names. With such model, we use Monte Carlo simulation to estimate the conversion rates of each potential customer in the future visiting. We believe our concept and the preliminary promising results in this paper enable the use of largely available customer online behaviours data for advanced digital marketing analysis.

\end{abstract}

\section{Introduction}

In the digital marketing world, one of the core applications is to analysis the customer online behaviours for creating more adaptive advertisements. Such applications rely heavily  on the online customer data analysis, in order to optimise the online ads return on investment. For instance, showing a car insurance promotion for the people that are looking for their house insurance is probably not appealing for these potential clients. Lack of customer behaviours understanding, the online advertising would be much less efficient. 
Indeed, for modern large companies, a slight optimization on the advertising strategies would result in millions of Euros for saving in their marketing budgets. 

Nevertheless, the most advanced analytic tools, such as Google Analytics \cite{ga}, can only provide analysis on customer visit statistic such as number of visits, page viewed and time spent on each page. These data are then used by expertises for analysis and interpretation, in order to derive an optimal marketing strategy.  Such traditional procedure of digital analysis needs to be done into a more automatic and intelligent way. 

Meanwhile, the recent advances in machine learning, especially in artificial neural networks, have shown very promising results in automatically learning the hidden patterns from the data \cite{lecun2015deep}. Over the last decade, we have seen the emerging applications using deep artificial neural network learning techniques in language processing \cite{blunsom2014convolutional,zhang2015character}, speech recognition \cite{Heiga}, video sequence analysis \cite{donahue2015long}, time series predictions \cite{xingjian2015convolutional} or  even modelling consumer click information on website for recommendation \cite{hidasi2015session}. These applications have attracted the machine learning researchers. More recently, the successful applications of Neural networks to these problems have shown very promising results and encourage more and more researchers to discover more possibilities to apply neural networks to similar problems. 

In addition, neural networks, especially deep neural networks, have been shown empirically improving their performances when the number of training data increases \cite{lecun2015deep}. Their performances compared to conventional machine learning algorithms make it possible to enable applications in large size training situation such as image classification, language translation. In digital marketing world, a huge amount of data can be easily collected. This propriety enable the new applications in recommended system \cite{hidasi2015session,smirnova2017contextual}, online advertising \cite{mcmahan2013ad,Wang2017}. 

In this paper, we are interested in applying neural networks for modelling the customer online visit behaviours, in order to estimate the probabilities for conversion rates.  More technically, we use Convolutional Neural Networks for the user searched keywords and the visited page names embedding, the embedded feature vectors are then fed into  Recurrent Neural Networks for modelling the in-session customer visited journey. For prediction, we use the Monte Carlo strategy to simulate n-steps forecasting in order to estimate the probabilities of online conversion for each predefined objective.

\section{Related work}

Neural networks have been proven to be effective for accomplishing difficult tasks in various domains \cite{lecun2015deep}. Among the most famous architectures, Convolutional Neural Networks (CNNs) and Recurrent Neural Networks (RNNs) have received increasing attention in recent years, thanks to their success when dealing with image data and sequential data such as language, speech and music.

In case of CNNs, they are mostly used in 2D array of data in the image processing domain \cite{lecun2015deep}. Thanks to the concept of using convolution layer that extracts the locally connected neighbourhood values (\textit{e.g.} a patch in image) in a non-linear way, and of using a pooling layer which allows abstracting the information. Inside a Deep CNN architecture, the output of previous layer can feed the reduced size image into next convolution layers that helps to extract more abstract high-level features successively. 
CNNs have also been used for natural language modelling \cite{blunsom2014convolutional,zhang2015character} for the similar reasons. In language processing tasks, the convolution applies in one-dimension that could extract pertinent features from words or n-gram given the input sentence. 

RNNs, instead, are dedicated for sequential data. They are used for learning the long term dependencies with the hidden unites that encode all historical information. For example, RNNs have been applied for machine translation problems with an encoder that models the whole sentence into a fixed length vector representation \cite{sutskever2014sequence}, in such way that similar semantic meaning sentences would be closer in the representation space. Thanks to their dependencies modelling capability, they have also been found to be very powerful for predicting the next characters, next words or next phonetics. Using the next steps prediction, lots of interesting applications, such as music generation \cite{jozefowicz2015empirical}, language translation \cite{sutskever2014sequence} and speech recognition \cite{graves2013speech} have been developed. These applications have achieved an excellent performance, allowing them to be industrialised and used in the real world environment.

More recently, the trends of combing the neural networks in order to accomplish more complex tasks has emerged. For example, image automatic caption problems have been attempted in  \cite{vinyals2015show} with a CNN that takes input as image and a RNN that generates the caption language. Video sequence classification problems seem also benefit from combining different architectures \cite{malinowski2015ask,pei2017temporal}: CNN for individual image frame processing and RNN for encoding the long-term contextual information.  Similarly, such strategy of local and global information taken into account by CNN and RNN has been used in Natural Language Processing problem such as Named Entity Recognition in \cite{chiu2015named}. 

In this paper, we benefit of advantages of both CNN and RNN for modelling the customer online visit behaviours, which will be detailed in the next sections. 

\section{Neural network for customer online behaviours modelling and prediction}

Our objective is to construct a model that can simulate the online visit behaviours of customers. The model would be used for targeting (re-targeting) the potential customers that might finish a conversion later on the site. To achieve such goal, the model should be capable of modelling the customer journey in a effective way, and be able to generate similar customer online journey in such way that proximate the real behaviours of visitors on the site. 

To do so, we rely on three key components: a CNN for embedding the searched keywords and page names, as they provide critical information about the visitor's motivation; a RNN for modelling the long term dependencies among different pages thanks to  its capacities of modelling sequential data; and Monte Carlo process for simulating the customer journey in order to estimate the probabilities of conversion rate for each predefined objective.

\subsection{CNN for keywords and page name embedding}

\begin{figure}[!ht]
	\centering
	\includegraphics[width=8cm]{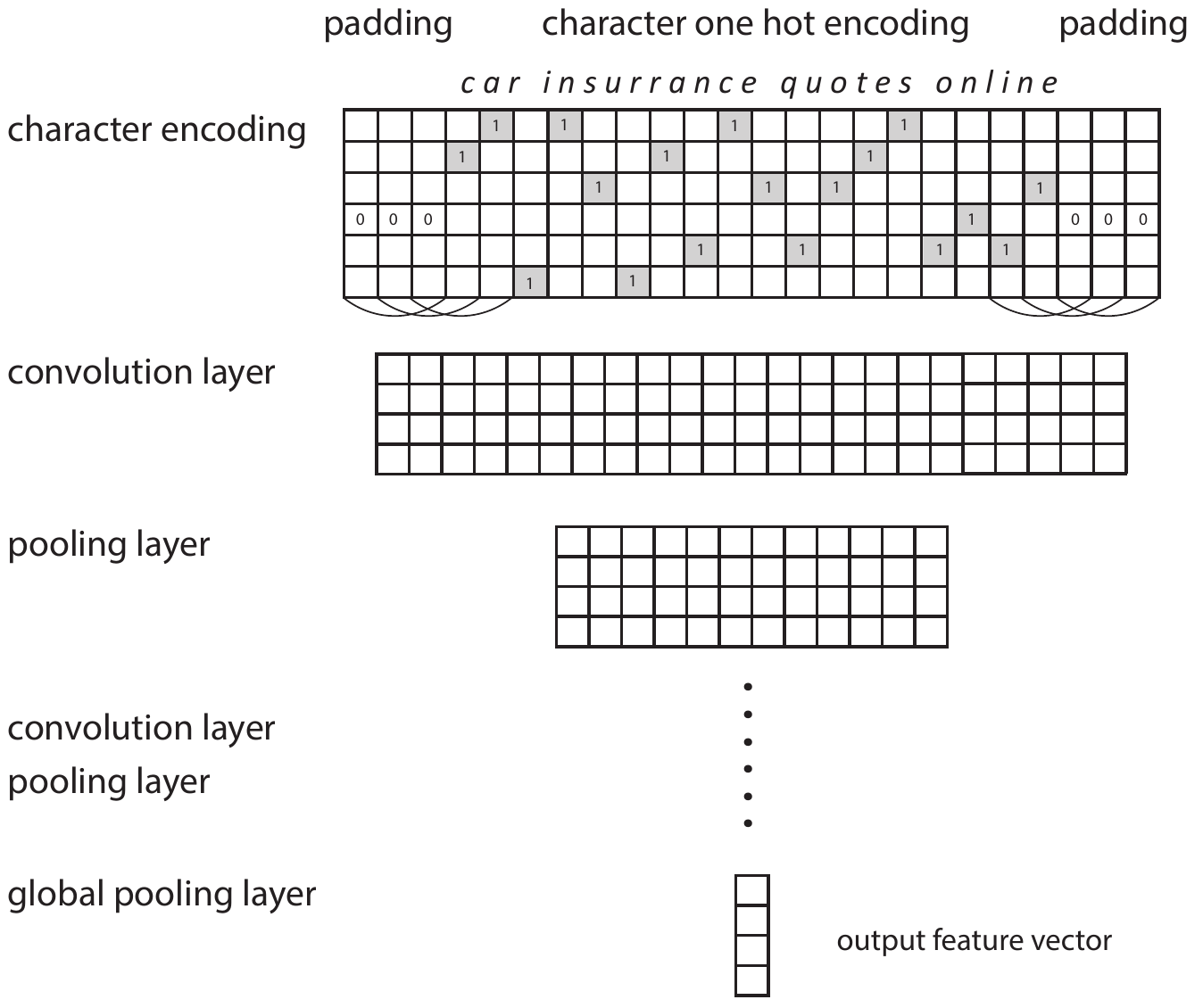}
	\caption{CNN architecture for embedding the texts (searched keywords or page names). The model takes input of one phrase and outputs a fixed dimension vector that encodes the character level features of the phrase, which will be used later as the input for RNN.}
	\label{fig:cnn}
\end{figure}

CNN has shown its capacity of modelling the natural language \cite{zhang2015character}. In our context, CNN is used to encode the keywords that user searched in Google Search Engine, and the page names in the site (predefined by the site owner). These two types of texts need to be carefully captured as both give the model crucial information about the next steps of customer visits. For examples, customers asked in search engine for giving a quote for their car insurance would probably go through the procedure of quote estimation in their visiting journey, or people who have visited an appointment page in the site would probably guided to the confirmation pages.

The CNN, as shown in Fig.\,\ref{fig:cnn}, allows us to extract character level features from each phrase. Such character level features gives us the possibility to model various type of word, special characters abbreviation used in page naming, or typos in user searched keywords. The key components in the CNN is composed by convolution layer, Max pooling layer, where deeper architecture (by repeatedly set the convolution and Max pooling layers on top of each other) allows for more abstract high level feature extraction.

\subsection{RNN for customer visiting journey modelling}

\begin{figure}[t]
	\centering
	\includegraphics[width=\textwidth]{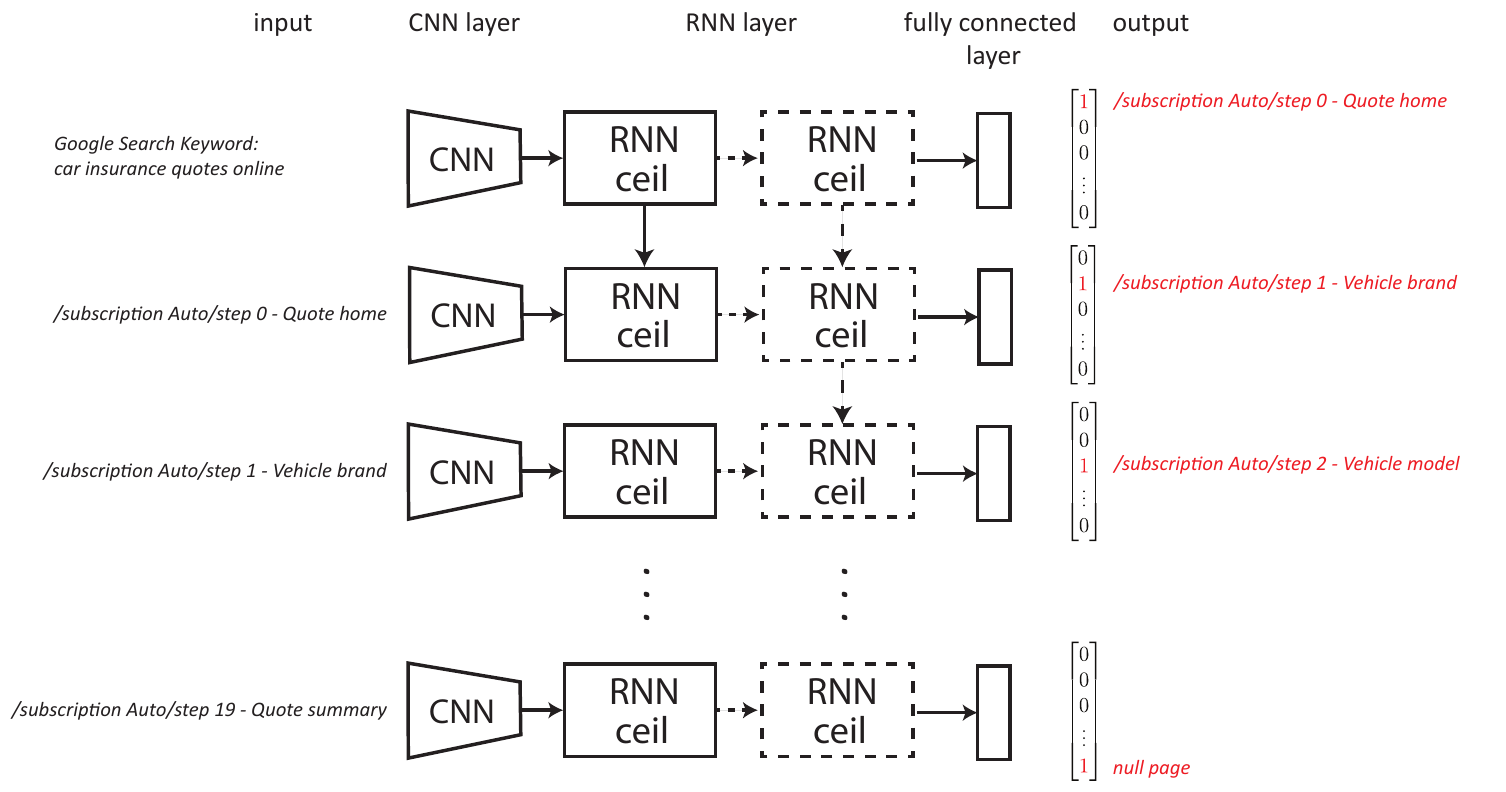}
	\caption{RNN architecture for modelling the customer journey. The input at each step is the embedded CNN feature vector of keywords or page names, and the output is one-hot encoded succeeding page.}
	\label{fig:lstm}
\end{figure}

In order to learn the dependencies among each step of costumer visited journey, we use multi-layer LSTM \cite{sutskever2014sequence} as our recurrent architecture shown in the Fig.\,\ref{fig:lstm}. The input of RNN is the feature vector that extracted from CNN layer using the user's search keywords in Google Search Engine, or url page names of the site. The output of RNN is, coped with fully connected layers, one-hot encoded page name (among a list  of all possible pages ${c_1, c_2, \ldots, c_N}$). 

The loss function of our recurrent network is computed as the sum of cross-entropy between the prediction at each step $t$ and the true visited page at each step $t+1$. In other words, we use the label as $y^t = x^{t+1}$ to train our network, with the loss function written as: 

\begin{equation}
L = \sum_{t=1}^T L^t = -   \sum_{t=1}^T \sum_{i = c_1}^{c_N}  y^{t}_i \log(\hat{y}^{t}_i) = - \sum_{t=1}^T  \sum_{i = c_1}^{c_N}  x^{t+1}_i \log(\hat{y}^{t}_i)
\end{equation}

In addition to this standard loss computation, we incorporate the dwell time (time spent on visiting the page) of each action, as it is highly related to the customer interests in certain pages \cite{yi2014beyond}. To do so, we duplicate the action for several step by a factor which relative to the dwell time. By incorporate such action,  the final construct visited pages will be closed to the real customer journey. 

It should be noted here that we also add one special page named ``null page'' indicating that the customer finished the visit and quit the site. Such choice is imperative as it reflects the reality that a user might quit the site at certain moments. More importantly, it indicates the end when running a simulation which described in details in the next section.

\subsection{Monte Carlo for customer journey simulation}

Here, we considered a problem of multi-steps forecasting knowing passed states $U$. The objective is to estimate the customer conversion rates knowing their visited pages. This corresponds to estimate a temporal evolution of probabilities distribution for each actions $p(s_{t})$. 

\begin{equation}
p(s_{t}|U) = \sum_{s_{t-1} = c_1}^{c_N} \cdots  \sum_{s_2 = c_1}^{c_N}  \sum_{s_{1} = c_1}^{c_N} p(s_{t} |s_{t-1},s_{t-2},\ldots, s_1,U) \cdots p(s_{2} |s_{1},U ) p(s_{1}|U)
\label{eq:mc} 
\end{equation}

As one may see in Eq.\,\ref{eq:mc}, such high dimensional probabilities calculation prevent us from performing explicit computation. We thus propose to use Monte Carlo simulation to estimate $p(s_{t})$ in order to calculate the user conversion. 

\begin{figure}[t]
	\centering
	\includegraphics[width=\textwidth]{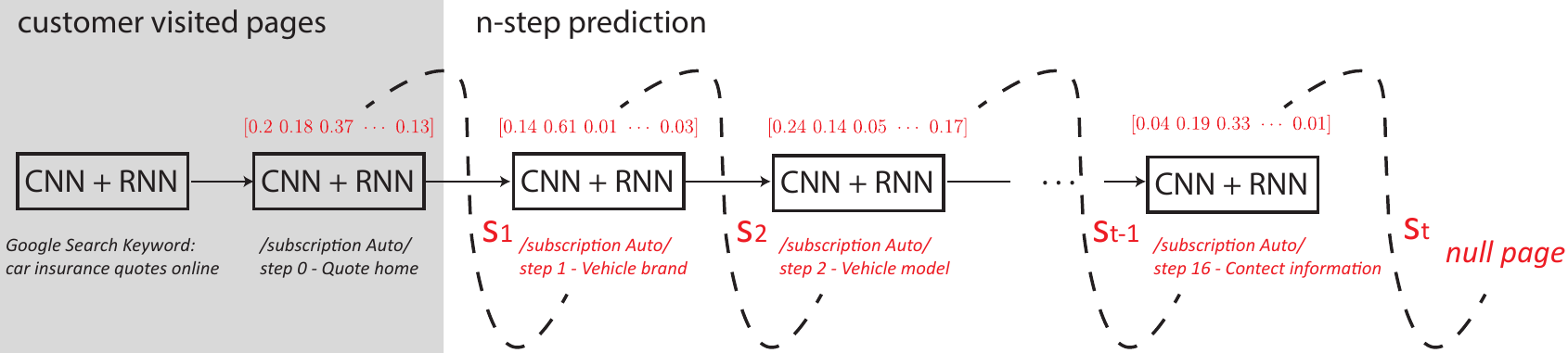}
	\caption{Monte Carlo simulation of complete customer journey given their visited pages. A importance sampling is done at each step $t$ according to the prediction probability distribution,  and the selected page $s_t$ will be feed into network in order to generate next prediction at step $t+1$.}
	\label{fig:mc}
\end{figure}

The procedure is shown in Fig.\,\ref{fig:mc}, where we use an importance sampling technique relying on the predicted probability at each step $t$, then use the sampled page name to feed into our neural network at next step $t+1$. According the law of large numbers, the estimation will be approximate to the real distribution given a large number of sampling instances.   Note that sampling method in RNN is quite often used in the text generation tasks \cite{vinyals2015show}, where the predictions $s_t$ are chosen as, instead of randomly, maximum probability at each step to given the most probable sentence.

\section{Results}

\subsection*{Experimental setup}

In order to evaluate the effectiveness of proposed neural network architectures, we use a subset of one month online visit logs at one French insurance company website \url{www.gmf.fr}, which corresponds 10 000 random selected records. We split further 80\% for training and 20\% for evaluations. The reported losses and accuracies are calculated by averaging the all the next page predictions (for each line of sequential record, we have $n$ predictions if customer visited $n$ pages) using previous pages, similar as in Eq.\,\ref{eq:mc}. 

For the neural network architecture, we use LSTM ceil as our sequential RNN model. We compare the performance of one-hot encoding for page name to a CNN model for user keywords and page name embedding. All the hyperparameters, \textit{i.e.} number of CNN and LSTM layers, convolution kernel size, number of parameters in the hidden layers, are tuned with best performance on evaluation set.  

\subsection*{The results of modelling}

We show the preliminary results of 3 different combinations of neural networks architectures. LSTM use the one-hot encoded page name as input, while CNN + LSTM use CNN for embedding user searched keywords and page names, as illustrate in Fig.\,\ref{fig:lstm}. In the end, ensemble CNN + LSTM uses 5 parallel identical architectures for modelling user behaviours. The reported results have been calculated with 2 CNN layers (2 combinations of convolution layer with $64$ filters and max pooling layer with size $4$ as in Fig.\,\ref{fig:cnn}), and 2 LSTM layers with $128$ hidden units, 1 fully connected layer with $256$ hidden units and 1 softmax layer with $568$ classes (different frequently visited pages on the site). We also apply Dropout techniques with $0.5$ drop rate at fully connected layer.

\begin{figure}
	\begin{minipage}[t]{0.55\linewidth}
		\centering
		\includegraphics[width=1\textwidth]{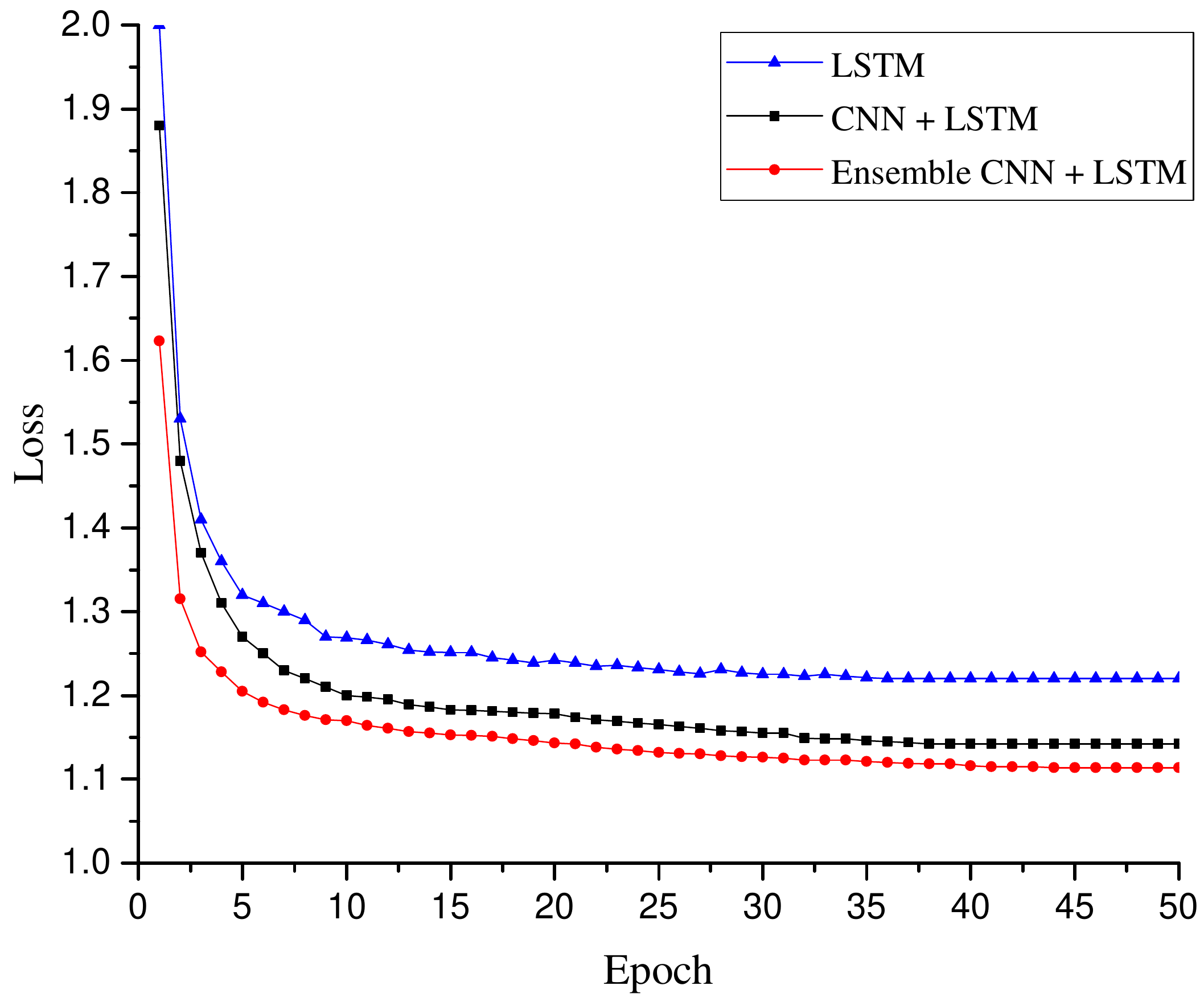}
		\captionof{figure}{Cross-entropy loss \textit{w.r.t.} training epoch}
		\label{fig:image}
	\end{minipage}
	\begin{minipage}[c]{0.38\linewidth}
		\vspace{-4.3em}
		\centering
		\begin{tabular}{ l c }
			\toprule
			Methods   & Accuracy  \\
			\midrule
			LSTM  & 63.5 \\
			CNN + LSTM & 65.3  \\
			Ensemble CNN + LSTM & 65.8  \\
			\bottomrule
		\end{tabular}
	    \vspace{2.2em}
		\captionof{table}{Prediction results on validation set}	
		\label{tab:results}
	\end{minipage}
\end{figure}

We can see in Fig.\,\ref{fig:image} and Tab.\,\ref{tab:results}, LSTM can model, at certain levels, the complex online behaviours, yielding a $63.5\%$ accuracy for step-by-step page predictions. Thanks to the CNN layers for capturing the semantic meaning of user searched keywords and page names, we obtain a $1.8\%$ improvement. The accuracy can be further improved by using ensemble technique, which by averaging the predictions results of each neural network, gives us the best result of $65.8\%$. It should be noted here that the randomness in online visit behaviours can not be ignored.  Even with $65.8\%$ accuracy, the model can be capture some interesting behavioural patterns, detailed in the next session.

\subsection*{The results of simulation}
							  
\begin{figure}[t]
	\centering
	\includegraphics[width=0.45\textwidth]{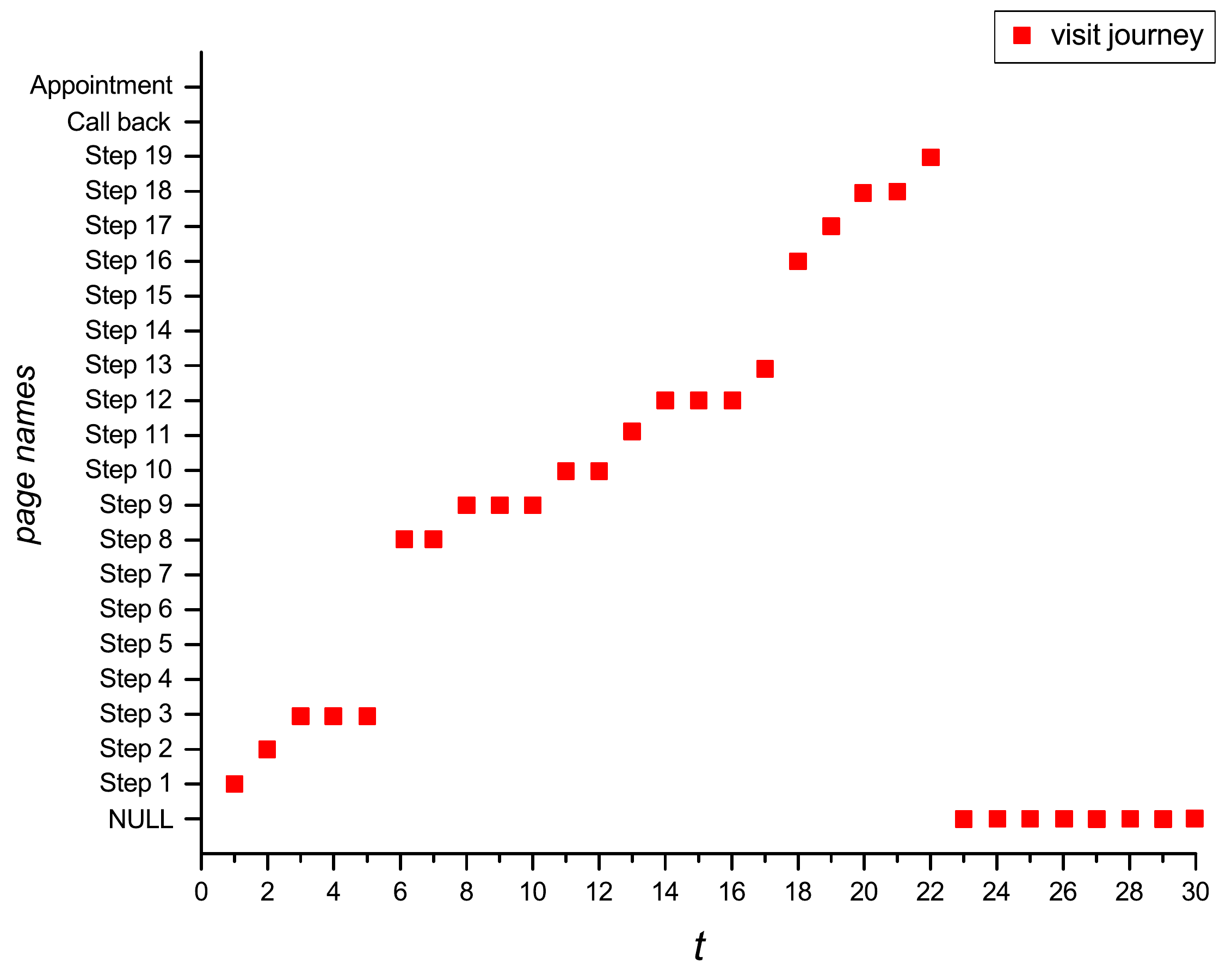}
	\quad
	\includegraphics[width=0.45\textwidth]{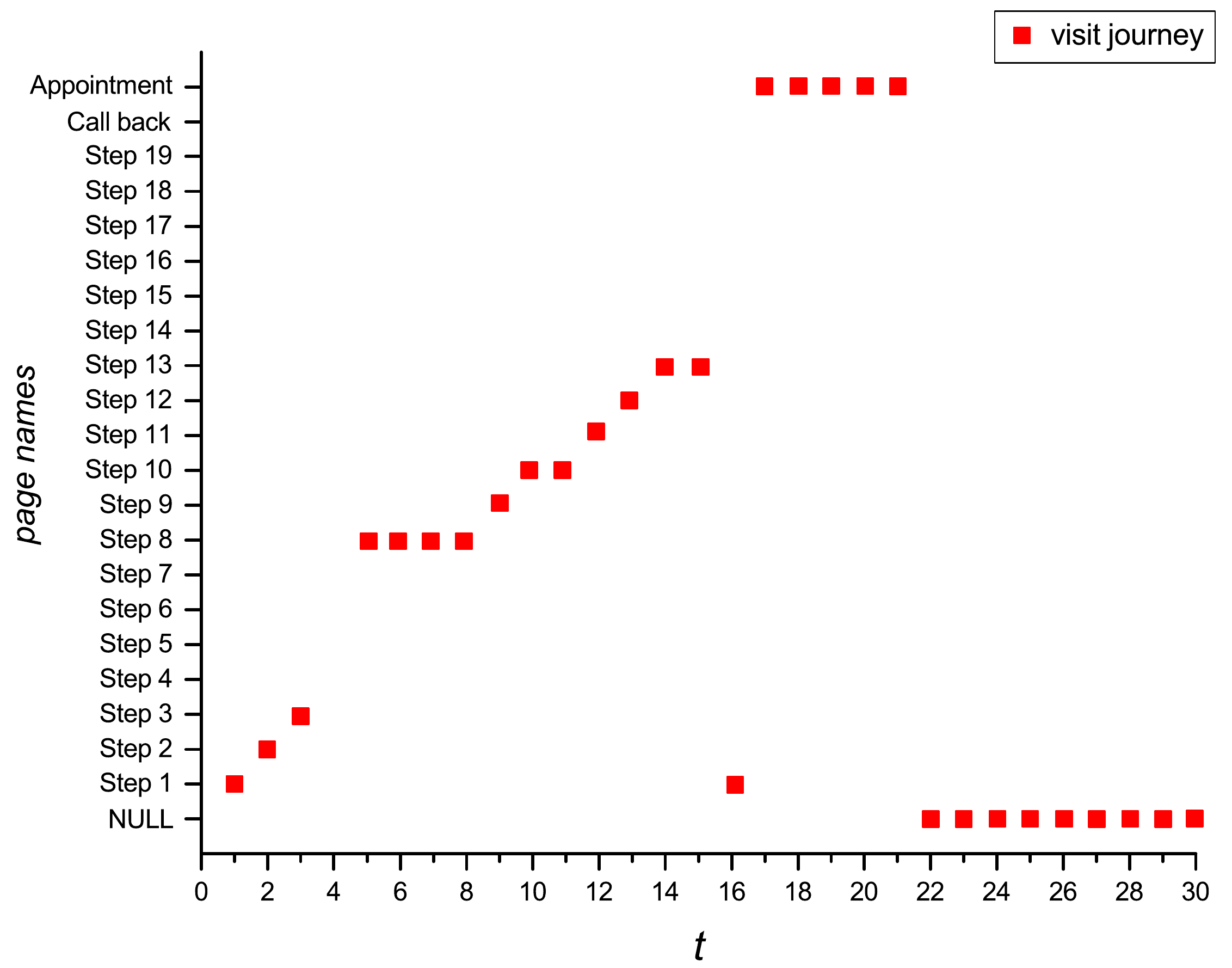}
	\caption{Two Monte Carlo simulations of customer visit journey for demanding car insurance quotes.}
	\label{fig:simu}
\end{figure}

After training the model with one month of logs of visit journey, the model is able to mimic the actual visit behaviours. In Fig.\,\ref{fig:simu}, we illustrate two simulations that are generated with keywords as ``car insurance quotes online'' and customer's first visited landing page. Note that these long-range steps (30 steps) predictions are generated using only two steps of information at the beginning of visit.  

We can see the first simulation represents one finished process of quote estimation from step 1 to step 19, where the missing steps are special pages that not often visited, and where the repeated steps are the ones require more times to fill the online form such as step 3 for the specific characteristics of their car, and step 12 and 13 for all the personal information related to car insurance. The second figure represents one situation that the customer is blocked at certain step and quite the procedure of online quote estimation, instead, took the appointment at agency. 

Both simulate situation are realist according to the expert in client management department. Therefore the generated simulation used for conversion rate estimation should also be realistic.

\section{Conclusion}

In this paper, we applied neural networks for modelling the customer online visit behaviours. As the visit logs are naturally organized in the form of sequence, we utilize the well studied sequential RNN to model them. In addition, we tried to capture the semantic meaning of user searched keywords  and page names for further improving the results.  
The preliminary results illustrate the capability of neural networks for modelling online visit sequence, gives the possibilities for predicting the future conversion rates with Monte Carlo simulations. Such conversion prediction is calculated for each customer and for each objective, enable the precise online advertising retargeting. The introduced method can make better use of largely available customer online behaviours data for advanced automatic digital marketing analysis. 
In the next steps, we will investigate the possibilities of bringing in advanced machine learning framework. Attention mechanism \cite{bahdanau2014neural} would be our first direction to explore for further improve the results. We also plan to integrate the proposed methods as utility estimation module in online display advertising \cite{cai2017real} within reinforcement learning framework for programmatic marketing.

\bibliographystyle{splncs03}
\bibliography{refs}

\end{document}